# A Hybrid CNN-LSTM deep Learning Model for Intrusion Detection in Smart Grid

Abdulhakim Alsaiari and Mohammad Ilyas

Department of Electrical Engineering and Computer Science, Florida Atlantic University, Boca Raton, FL, USA

*ABSTRACT*

*The evolution of the traditional power grid into the "smart grid" has resulted in a fundamental shift in energy management which allows the integration of renewable energy sources with modern communication technology. However, this interconnection has increased smart grids' vulnerability to attackers, which might result in privacy breaches, operational interruptions, and massive outages. The SCADA-based smart grid protocols are critical for realtime data collecting and control, but they are vulnerable to attacks like unauthorized access and denial of service (DoS). This research proposes a hybrid deep learning-based Intrusion Detection System (IDS) intended to improve the cybersecurity of smart grids. The suggested model takes advantage of Convolutional Neural Networks' (CNN) feature extraction capabilities as well as Long Short-Term Memory (LSTM) networks' temporal pattern recognition skills. DNP3 and IEC104 intrusion detection datasets are employed to train and test our CNN-LSTM model to recognize and classify the potential cyberthreats. Comparing to other deep learning approaches, the results demonstrate considerable improvements in accuracy, precision, recall, and F1-score, with a detection accuracy of 99.70%.*

## 1. INTRODUCTION

### 1.1. Smart Grid (SG)

A smart grid (SG) is a modern electrical system that employs digital communication, control technology, and automation to improve the efficiency, reliability, and sustainability of energy distribution and consumption. The integration of Information and Communications Technologies (ICT) in Smart Grids (SG) allows Internet of Things (IoT) devices to actively participate in power system operations, thus enhancing grid performance in various aspects, including real-time monitoring, forecasting, peak load estimation, immediate response, power factor enhancement, and fault detection and analysis [1]. Moreover, smart grids allow for a two-way transfer of information and power, as compared to conventional power grid systems that depend on a one-way flow of energy from generators to customers. This two-way functionality enables dynamic responses to fluctuations in demand, the integration of renewable energy sources, and accurate management of electricity distribution. To provide a secure, reliable, and sustainable electric power system, smart grids use advanced communication technologies, which allow for effective power management and communication [2]. However, integrating power systems with ICT technology exposes the grid to more sophisticated threats and rapidly advancing attack techniques, potentially resulting in significant financial and operational consequences. In addition, the growth of smart grid technologies poses substantial cybersecurity challenges because of the presence of insecure legacy systems such as Supervisory Control and Data Acquisition (SCADA), vulnerabilities in Transmission Control Protocol/Internet Protocol (TCP/IP), and new attack surfaces. Thus, Intrusion Detection Systems (IDSs) are essential for securing the smart grid infrastructure, which is a crucial concern that requires effective solutions to ensure data reliability, integrity, and confidentiality [3].

## 1.2. SCADA (Supervisory Control and Data Acquisition)

Modern industrial operations rely on Supervisory Control and Data Acquisition systems (SCADA) to monitor and control vital infrastructures like electric power grids, water and wastewater treatment facilities, petrochemical refineries, mining activities, and oil and gas pipelines [4]. SCADA is a crucial system for managing and operating modern smart grids, which automate electrical infrastructure and allow for remote control and monitoring in real time. It connects field equipment, such as sensors, meters, and circuit breakers, to centralized control centers via communication networks, allowing operators to effectively manage electricity generation, distribution, and consumption [5]. The five core components of a SCADA system are as follows: a) measurement instruments, b) logic controllers, c) a Master Terminal Unit (MTU), d) a communication interface, and e) a Human Machine Interface (HMI). Measurement instruments, including sensors and actuators, track environmental conditions and gather data such as temperature, voltage, and current. A logic controller is typically a Programmable Logic Controller (PLC) or remote terminal unit (RTU) that is responsible for the collection of measurements from preceding measurement instruments, the identification of operational abnormalities, and the activation/deactivation or configuration of other devices. The MTU functions as the central server or host, enabling operators to control and adjust the logic controllers primarily through a user-friendly interface known as the HMI. Finally, communication between the MTU and logic controllers is facilitated by a communication interface, including industrial protocols such as Distributed Network Protocol 3 (DNP3), security mechanisms like IDS, and hardware components such as switches, and data concentrators [6]. Furthermore, SCADA systems in electric networks are essential infrastructures composed of computer-based, networked systems that exchange critical data, but their reliance on information technology makes them vulnerable to intrusion attacks [7]. Therefore, evaluating system security by considering potential attacks from network intruders across communication networks is essential for maintaining the security and operational stability of modern electric infrastructure. Moreover, SCADA systems support demand response programs by dynamically balancing supply and demand, improving efficiency and service reliability. As power grids become more digital, SCADA systems are essential for improving performance, assuring cybersecurity, and allowing predictive maintenance, thus making the smart grid more robust, sustainable, and adaptable to future challenges [8] (see figure 1).

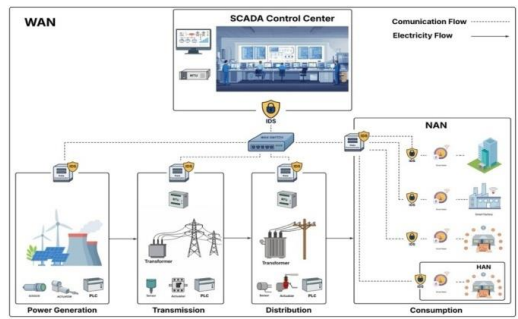

Figure 1. Smart grid architecture with integrated IDS and control components

### 1.3. SCADA Protocols

Different components of a smart grid, such as energy suppliers, utility companies, meters, and consumer side devices, can exchange information through various communication protocols. In a smart grid, different components have diverse requirements in terms of performance, security, power consumption, and reliability. Therefore, many functional activities within smart grids utilize specialized protocols that are more appropriate for their unique requirements [9]. Below is an overview of the most popular protocols employed by smart grids:

- DNP3

The electric utility sector widely utilizes Distributed Network Protocol 3 (DNP3) as a crucial communication protocol, particularly in smart grid applications. DNP3 enables communication between devices in SCADA (Supervisory Control and Data Acquisition) systems. This protocol links centralized control systems, such as SCADA, to remote terminal units and Intelligent Electronic Devices (IEDs), such as sensors and actuators, through various communication networks, thereby enhancing the monitoring and management of electrical systems. [10]. With their high level of intelligence, RTUs can communicate with SCADA masters using various protocols, such as DNP3 and Modbus. This guarantees the reliable and secure transmission and reception of sensing data and orders. Interconnected electronic devices (IEDs) include sensors, actuators, and programmed logic controllers that are linked to the RTU, facilitating realtime data exchange and operational efficiency. In fact, electric utilities often use DNP3 for SCADA systems, especially in North America [11].

- IEC 61850

Currently, substation automation systems and communication protocols in smart grids are mostly based on the IEC 61850 standard. These substations are linked by industrial Ethernet switches, allowing intelligent electronic devices and automation applications to communicate via the substation LAN using the IEC 61850 standard. Moreover, IEC 61850 is a crucial communication protocol for contemporary and future smart grids because it allows interoperability among devices from various vendors and ensures real-time communication, high-speed protection, monitoring, and control of power systems within substations and throughout the grid. GOOSE (Generic Object-Oriented Substation Event) and SMV (Sampled Measured Values) are two critical protocols offered by the IEC 61850 standard for transmitting crucial communications. These protocols facilitate the broadcasting of multimedia messages over the LAN and utilize a switched Ethernet network for communication. While GOOSE facilitates rapid, event-driven communication to execute protection functions within milliseconds, SMV enables the real-time transfer of measurement values such as current and voltage from sensors to IEDs [12]. However, IEC 61850 lacks security features, which implies it is unable to ensure confidentiality, integrity, and authentication in network communications. Therefore, the evolution of IEC 62351 aims to improve the security of IEC 61850 communication.

- Modbus

Modbus is a popular industrial communication protocol, notably for SCADA systems in the energy sector. Its simplicity, flexibility, and ease of implementation make it excellent for applications that require consistent communication between field devices and control systems, especially for local automation, monitoring, and DER integration in the energy and power fields. The general Modbus frame is referred to as the Application Data Unit (ADU), which consists of three primary sections: Protocol Data Unit (PDU), Addressing, and Error Checking. The PDU contains the essential information about Modbus packets, comprising the function code and corresponding data. Each function code delineates a distinct functionality. The addressing and error-checking functions depend on the Modbus version (e.g., RTU vs. TCP/IP). While each master-slave pair is identified by unique IDs to recognize devices, error checking mechanisms are conducted using Cyclic Redundancy Check (CRC) to guarantee data integrity [13]. Simplicity, interoperability, and ease of integration of the Modbus protocol render it appropriate for small-scale implementations, including DER monitoring, microgrid management, and power quality assessment. However, due to its security limitations and lack of real-time communications, more advanced protocols like IEC 61850 or DNP3 will oversee vital real-time operations and extensive automation [9].

- IEC 60870-5-104 (IEC 104)

Smart grids primarily use IEC 60870-5-104, also known as IEC 104, as a fundamental communication protocol for power system automation. IEC 104 allows for real-time data transmission, remote monitoring, and control. It guarantees that utilities can keep the system stable, incorporate renewable energy, and respond quickly to problems. By including support for TCP/IP networks, it improves the compatibility of the older IEC 60870-5-101 standard, making it more suitable for the efficient operation of SCADA systems in contemporary smart grid applications [11]. However, IEC-104 devices are susceptible to unauthorized connections, data manipulation attacks, and man-in-the-middle attacks [14]. Hence, to improve security, several contemporary grid systems integrate VPNs, firewalls, and network intrusion detection systems (IDS). Moreover, several grid operators are using IEC 62351 for secure authentication and encryption.

- IEC 62351

IEC 62351 is essential for enhancing the resilience of smart grids and various industrial control systems (ICS) against cybersecurity attacks. The technical committee of the International Electro technical Commission (IEC) created the IEC 62351 standard to ensure the security of communication protocols utilized in power system operations, including IEC 60870-5-104 (IEC 104), IEC 61850, and DNP3 [12]. Moreover, the security mechanisms of the IEC 62351 standard can be implemented at the application and transport layers of IEC 60870-5-104 and other protocols to protect the confidentiality, integrity, and availability of communication and control activities amidst the growing digitalization of grids. address cybersecurity concerns [14] As the energy landscape shifts toward greater renewable integration and decentralized systems, IEC 62351 will play an increasingly important role in sustaining confidence and stability.

- IEEE 2030.5 (Smart Energy Profile 2.0 or SEP 2.0)

IEEE 2030.5 Smart Energy Profile 2.0 (SEP2) is a communication protocol that is intended to enable the process of establishing communication between the smart infrastructure and consumers. It was initially introduced in 2016 as an IoT-based protocol by the Zigbee Alliance for applications that require interaction with demand response programs, electric vehicles (EVs), and Distributed Energy Resources (DERs). Subsequently, it is adopted by IEEE as IEEE 2030.5 with the purpose of facilitating the exchange of important data, including energy consumption, pricing, and demand response. The protocol became increasingly popular due to its ability to provide secure, interoperable communication between a variety of smart grid consumer devices and ecosystems [11]. Even though utilities are increasingly adopting distributed energy resources and electrifying transportation, SEP 2.0 will continue to improve grid stability and efficiency by coordinating flexible energy assets in real time. Furthermore, IEEE 2030.5 provides a framework that is both scalable and secure for energy management and consumer engagement within the context of the emerging smart grid environment. However, it is limited for real-time control in comparison to protocols such as IEC 61850.

### 1.4. Denial of Service (DoS)

A DoS attack occurs when attackers flood a network or individual devices with excessive or malicious traffic, exhausting their resources and rendering them unusable. This disturbance may harm system operations by delaying or preventing control signal communication, resulting in critical infrastructure failures. Smart grids are vulnerable to denial-of-service (DoS) attacks because they rely on continuous data exchange and communication amongst different components to ensure stable operation. The communication system of a smart grid includes SCADA systems, IEDs, RTUs, and PLCs, all of which necessitate real-time data for effective decision-making [15]. A DoS targeting these components can prevent the control center from obtaining correct data or sending control orders, thus disrupting activities such as load management, fault detection, and power balancing. In a smart grid, the power system's condition changes in real time, making it difficult to simulate with high accuracy. Grid operators use dynamic state estimation as a critical procedure to monitor the system, heavily relying on real-time measurement data to maintain accuracy. When these measurements lack or are delayed, the accuracy of the estimation suffers dramatically. For example, if communication from a substation is disrupted, operators may lose access to key data. This blind zone could prevent them from rapidly identifying potential defects, raising the possibility of power outages or equipment damage. Reliable connectivity and accurate data are critical for keeping the grid stable and efficient. DoS attacks also target security vulnerabilities in common communication protocols used in smart grids, such as DNP3, Modbus, and IEC 61850. Attackers can exploit these protocols due to their inadequate security design [16]. Furthermore, DoS attacks disrupt lawful communication and degrade system performance by flooding the communication network with excessive packets. For instance, Distributed Energy Resources (DERs), such as solar and wind power facilities, rely on network connectivity to coordinate with the grid. DoS attacks can cause loss of control or disconnections, disrupting power supplies. The consequences include grid instability, outages, delayed reaction times, and the possibility of cascading failures, which might cause extensive electrical outages or even physical infrastructure damage [17]. In fact, while smart grids are continually incorporating more digital devices and Internet of Things technologies, their attack surface is expanding, which makes them more susceptible to DoS attacks. To reduce potential risks caused by DoS attacks, utilities must implement a variety of preventative measures. Intrusion Detection Systems (IDS) are critical for identifying unusual traffic patterns and alerting operators to possible DoS attacks. Traffic filtering, firewalls, and rate-limiting methods can assist in preventing malicious traffic from overwhelming the network. Furthermore, using multiple communication pathways guarantees that key processes are still available even if one channel is compromised. In addition, adopting safe protocols such as IEC 62351, which improve the security of SCADA communications, increases the grid's defenses [15]. Nonetheless, incorporating new connections to essential infrastructure heightens the potential for cyber threats such as DoS. Consequently, cybersecurity research has increasingly focused on intrusion detection systems (IDS) in SCADA networks to safeguard smart grids against potential attacks. [8].

### 1.5. Intrusion Detection Systems (IDS)

Data intrusion attacks rank among the most common cyber threats to the security of smart grids. These attacks typically categorize into three primary types: false data injection (FDI) attacks, load redistribution (LR) attacks, and denial of service (DoS) attacks. Cyber attackers leverage these methods to manipulate and alter the data involved in the smart grid operations, aiming to disrupt grid stability, gain financial benefits, or even cause physical damage to grid infrastructure [18]. The effective detection and separation of abnormal data from normal data are crucial roles of contemporary Intrusion Detection Systems (IDS), which play a vital part in overseeing smart grid security and maintaining situational awareness. IDS is essential for detecting attacks both prior to and after unauthorized access to the network has been achieved, ensuring data availability while protecting the system's confidentiality and integrity against potential threats [7]. In addition, the purpose of an Intrusion Detection System (IDS) is to monitor network traffic and analyze activity patterns, allowing them to detect and react to security breaches in real time. These systems are critical for ensuring the security and integrity of computer networks by identifying security threats such as denial-of-service (DoS) assaults, port scans, and injections of malware. Moreover, it can be difficult for traditional IDS methods to detect novel or advanced attacks as they are often based on previously established rules and signatures to identify common attack patterns [19]. In fact, as the smart grid becomes more digital and interconnected with various technologies, the smart grid's communication network has become prone to cyberattacks, which can result in significant disasters that impact both consumers and utilities. Thus, advanced intrusion detection systems that employ advanced techniques such as deep learning (DL) are essential to address the limitations of traditional IDSs and to enhance the smart grid's ability to defend against evolving cyber threats [7]. Deep learning algorithms demonstrated impressive abilities in automatically identifying intricate patterns and features from complex data, such as network traffic. These advanced IDSs can proficiently recognize unusual behavior and uncover emerging threats in real time. By providing better accuracy, scalability, and adaptability, advanced IDS is a promising solution for improving network security and addressing the risks caused by evolving cyber threats such as DoS attacks [19].

The integration of digital communication and automation technologies has led to the rapid development of smart grids, which represent the fundamental evolution of traditional power systems. While this development enhances sustainability, efficiency, and reliability, it also introduces significant cybersecurity challenges, especially to the communication infrastructure that supports Supervisory Control and Data Acquisition (SCADA) systems. These systems rely on protocols such as DNP3 and IEC 60870-5-104, which are increasingly vulnerable to modern cyberattacks [45]. The primary goal of this project is to create a robust and efficient Intrusion Detection System (IDS) for smart grid systems by leveraging deep learning (DL) capabilities. To accomplish this, the following contributions are made:

- We propose a hybrid deep learning model that combines Convolutional Neural Networks (CNN) and Long Short-Term Memory (LSTM) networks to capture both spatial and temporal features of network traffic, thus facilitating the detection of abnormal behavior in smart grid communication.
- For binary-class classification purposes, the model is trained and tested on DNP3 and IEC 60870-5104 intrusion detection datasets. These datasets reflect real-world smart grid cyberattacks, focusing on unauthorized commands and DoS threats against DNP3 and IEC 60870-5-104 protocols.
- Furthermore, extensive experimentations, including hyperparameter tuning, were conducted to evaluate and ensure the efficiency of the proposed CNN-LSTM model for IDS of smart grids.
- The results demonstrate remarkable improvements across key performance metrics such as accuracy, precision, recall, and F1 score, achieving an accuracy of 99.70% and nearly 100% detection rate.
- This paper is organized as follows: Section 2 provides an overview of relevant works in the field of smart grid cybersecurity and deep learning-based intrusion detection systems (IDS). Section 3 highlights the background and fundamental components of the deep learning architectures and implementation. Section 4 describes the architecture of the proposed CNN-LSTM model. Section 5 presents the employed datasets and covers the evaluation metrics, experimental setups, and results analysis. Finally, Section 6 concludes the research paper and outlines directions for future research.

## 2. RELATED STUDIES

Sophisticated cyberattacks are driving the rapid evolution of intrusion detection systems (IDS). Modern IDS systems use deep learning architecture, especially hybrid models combining neural network approaches instead of signature-based detection and manual feature engineering. Focusing on CNN-LSTM hybrid architectures, this section addresses deep learning-based IDS technology innovations, challenges, and future viewpoints.

Deep learning architectures have replaced signature-based intrusion detection systems with hybrid models that are doing exceptionally well. The authors of [20] demonstrated 99.7% accuracy with 45,000 parameters using a hybrid CNN-LSTM architecture, which included a 1D CNN with two convolutional layers (64 filters each) for initial feature extraction, and an LSTM component with 64 units and a 0.5 dropout rate. This architecture is suitable for real-time detection due to its parameter utilization and 1D CNN's simplicity. However, the researchers in [16] proposed a more complex hybrid architecture that incorporates squeezeand-excitation (SE) blocks, autoregressive (AR) components, L2 regularization, dropout layers with a rate of 0.5, and dynamic channel importance adjustment to enhance robustness against various attack levels. The SE blocks' adjustable channel weighting enhances feature discrimination but complicates processing. Through selective feature emphasis, CNN-LSTM architectures with SE blocks may increase model flexibility and computational efficiency.

The difficulty of network security feature extraction and selection drives deep learning-based IDS architecture improvements. On the CICDDoS2019 dataset, the research paper [21] suggested an intelligent agent-based system with automatic feature extraction and selection that outperformed existing methods by 99.7%. Their two-phase detection approach allowed dynamic reconstruction of detector agents based on optimal feature sets. In [22], multilayer LSTM networks are used to analyze time-series electric waveform data with only one voltage and one current sensor at the point of common coupling and offline training and online testing strategies to achieve superior performance with 0.05s window sizes. [23] showed how innovative activation functions like PRetanh might optimize recurrent neural networks for greater learning rates without divergence and steady performance. These studies demonstrate the significance of automated feature extraction in IDS development, highlighting the need to strike a balance between feature comprehensiveness and computational efficiency.

Real-time detection and computational overhead minimization are crucial in current IDS solutions. A deep CNN model was developed by [19], which used CNN layers with 128 and 256 filter units, GPU acceleration, and L1 and L2 regularization to achieve 99.79–100% accuracy across several datasets [19]. In comparison, [24] suggested the DDoSNet architecture, which used RNN-autoencoder design, unsupervised pre-training, and supervised fine-tuning to achieve 99% accuracy, superior precision, and recall in a four-layer encoder-decoder structure (64, 32, 16, 8 channels). Further, through neural network implementation, researchers in [25] showed how theory-guided and physics-informed techniques might improve model performance beyond typical training paradigms. Therefore, the balance between detecting accuracy and processing economy remains challenging, although hybrid designs are improving.

IDS development struggles with skewed datasets and unusual attack pattern detection. SMOTE modification is used in [3] to resolve dataset imbalance in the NSL-KDD dataset with 125,973 training samples across 42 features, concentrating on underrepresented attack classes. Taking a different architectural approach, [6] developed the DIDEROT system, which uses a dual-layer detection method to detect anomalies using supervised ML (0.997 accuracy) and an unsupervised autoencoder deep neural network (0.951 accuracy). Adding to these findings, a multilayer feed-forward neural network classifier was tested in [26], achieving 98.99% accuracy with a 0.56% false alarm rate utilizing just the top 20% significant features. This means combining data transformation methods with well-defined architecture approaches may help handle dataset imbalance while retaining detection efficacy.

Integrated temporal and spatial feature analysis is key to improving IDS capabilities. The research paper [27] showed that LSTM's memory cell design handles temporal dependencies better than CNN's hierarchical feature learning for spatial feature extraction. Also, in [28], independent recurrent neural networks (IndRNN) research allowed sequence processing across 5000-time steps and network levels up to 21 layers using new gradient management methods. Further, [29] used LSTM and GRU in a multilayer bidirectional configuration to show how architectural improvements might improve forward and backward temporal dependence capture. From these findings, IDS development has advanced to sophisticated temporal-spatial feature estimation, yet model interpretability must be maintained.

Existing and future IDS development challenges concentrate on evolving threat environments while retaining system efficiency. The study in [30] showed that signature-based techniques are limited and recommended hybrid designs that combine accuracy with anomaly detection flexibility. The authors of [31] achieved 97% AUC with thorough feature extraction across time and frequency domains using meticulous feature engineering and Bayesian hyperparameter adjustment. [17]'s study underscored the necessity for scalable algorithms that maintain efficacy without disrupting network operations and demonstrated that flow-based analysis generally outperforms deep packet inspection in large-scale networks. Advanced hybrid architectures that balance detection accuracy, computing efficiency, and threat adaptability while maintaining deployment practicality are the future of IDS development.

Deep learning-based IDS has made outstanding strides in cybersecurity. Hybrid architectures, especially CNN-LSTM ones, balance detection accuracy and computing efficiency well. Future research must address real-time detection, dataset imbalance, and model interpretability while designing more flexible and efficient architectures as threat environments change.
Convolutional Neural Networks (CNNs) are a type of deep learning architecture that utilizes convolutional operations to extract spatial features from data, making them extremely efficient in applications such as image recognition, anomaly detection, and network intrusion detection. Long Short-Term Memory (LSTM) networks are a particular type of Recurrent Neural Network (RNN) that can capture long-term temporal dependencies using gated memory cells, making them suited for processing sequential or time-series data. CNNs have been extensively utilized in tasks that require spatial feature extraction, while LSTMs are particularly efficient at sequence modeling. Their combination allows for the capturing of both temporal and spatial patterns. For example, [46] utilized a deep neural network framework integrating these architectures to inventory management performance in chaotic complexity systems, illustrating their capability to deal with highly dynamic and nonlinear environments. This demonstrated adaptability across multiple application domains emphasizes their suitability for addressing the spatial and temporal complexity inherent in smart grid intrusion detection.

## 3. BACKGROUND

Deep learning is a subset of machine learning that employs deep neural networks (DNNs), a form of artificial neural network (ANN) with multiple layers of correlated artificial neurons.

These networks are designed to learn hierarchical representations from large datasets, progressively transforming input data as it moves through layers. While lower layers capture simple features such as edges in an image, higher layers combine these features to recognize complex patterns, such as objects or faces [32]. The ability to automatically extract features from raw data, known as feature learning, separates DNNs from traditional ANNs. Furthermore, deep neural networks, with their complex topologies and sophisticated training techniques, allow for higher degrees of data abstraction and complexity. Networks that consist of more than three layers, incorporating at least two hidden layers, are defined as "deep," whereas networks with fewer layers are classified as shallow. The hierarchical structure of DNNs, which typically contains an input layer, several hidden layers, and an output layer, allows for a compact and efficient representation of input-output relationships [33].

Deep learning has garnered considerable attention and research initiatives due to its adaptability, efficiency, and ability to analyze vast data and identify complex patterns. Researchers are actively developing novel deep-learning approaches to address issues such as model interpretability, training with limited labeled data, and enhancing algorithm efficiency. A variety of deep-learning architectures have been designed to deal with different types of data and tasks. For example, Deep Belief Networks (DBNs) are generative models capable of unsupervised learning, Recurrent Neural Networks (RNNs) are particularly adept at handling sequential data, and Convolutional Neural Networks (CNNs) have shown remarkable success with structured grid-like data, such as images and audio [34]. However, the expanding diversity of DL architectures, as well as the steep learning curve required to understand them, present challenges for researchers and practitioners to adopt and implement them successfully. These challenges include developing strategies to train models effectively with limited labeled data, improving the efficiency of algorithms to make them faster and more resource-efficient, and enhancing models to make their decisionmaking processes more comprehensible and accurate. Despite these challenges, deep learning continues to transform computational approaches and solve real-world issues, affecting the future of technology across various sectors [32].

Various deep-learning architectures like CNNs and LSTMs have been developed to tackle different data types and tasks. Here are the key differences between CNNs and LSTMs based on their characteristics and applications:

a. Architecture

- Input, convolutional, pooling, fully connected, and output layers are the components that distinguish a standard CNN from other architectures. By applying filters (kernels) to the input data, convolutional layers can capture local patterns and hierarchical structures. This is accomplished by gradually decreasing the amount of the input data while simultaneously increasing the number of feature maps [27].
- LSTM is an advanced type of Recurrent Neural Network (RNN) that incorporates memory cells along with input, forget, and output gates to effectively mitigate challenges like vanishing and exploding gradients [35].

b. Feature Learning

- CNNs are efficient in automatically learning and extracting hierarchical spatial characteristics from raw input data, such as edges, textures, and shapes. They accomplish translation invariance and robust spatial pattern detection by adding convolutional and pooling layers in succession [36].

International Journal of Artificial Intelligence and Applications (IJAIA), Vol.16, No.5, September 2025

- LSTMs are excellent at learning temporal patterns and capturing sequential dependencies in the data. Their memory cells and gating mechanisms allow them to maintain context across time, recording intricate dependencies and long-term relationships [37].

c. Use Cases

- CNNs are critical in computer vision applications such as picture classification, object identification, and segmentation, providing cutting-edge performance by successfully extracting spatial patterns. Their significance extends to a wide range of fields, including medical diagnostics, autonomous vehicles, and security systems [34].
- LSTMs are particularly effective with sequential data, which makes them well-suited for various applications, including natural language processing (such as language modeling and sentiment analysis), time series forecasting (like stock market prediction and anomaly detection), and audio processing (including speech recognition and music generation). Their ability to capture long-term dependencies allows for applications across various fields, such as healthcare, finance, and multimedia analysis.

d. Input Data Size and Format

- CNNs are typically designed for handling structured grid data, including 2D arrays (images) or 3D arrays (videos), often requiring inputs formatted as matrices or arrays. Methods such as resizing, padding, or converting raw data into image-like formats allow CNNs to deal with variable-sized inputs efficiently for applications like anomaly detection and image classification. These approaches guarantee alignment with CNN architectures for a variety of applications [38].
- LSTMs are designed to handle sequential data of different lengths, including time series, text, or audio sequences, which makes them exceptionally adaptable. They are efficient at capturing long-term dependencies by choosing to retain or forget information through their memory cells. To effectively process variable-length sequences, techniques such as padding, truncation, or specialized architectures are employed to ensure optimal performance and reduce reconstruction errors [39].

A hybrid CNN-LSTM deep learning approach exploits the combined strengths of CNNs and LSTMs to tackle complex challenges related to spatial and temporal data. In addition, CNNs perform well at extracting spatial features, including local patterns and high-dimensional relationships, from inputs such as images or sequential frames. On the other hand, LSTMs are specialized in learning temporal patterns and sequential dependencies to effectively capture temporal dependencies and long-term context [16]. By leveraging the advantages of these DL techniques, a CNN-LSTM hybrid approach can be efficient to solve classification problems across several domains. For instance, this combination proves to be useful for tasks like intrusion detection, where CNNs recognize patterns in network behaviors, and LSTMs examine anomalies over time [40]. These deep learning techniques or their hybrid models are adaptable to solve a variety of cybersecurity challenges like intrusion detection, malware identification, and cyber-attack prediction, such as DoS. Eventually, it is essential to understand the unique strengths of each technique to develop a CNN-LSTM architecture customized to certain tasks [27].

## 4. THE PROPOSED CNN-LSTM MODEL

The model architecture illustrated in Figure 1 includes three CNN blocks and two LSTM blocks to improve the network's efficacy. The convolutional layers generate hierarchical feature maps by applying convolutional filters (kernels) to the input data, thereby extracting significant spatial features. In the convolutional kernel, the filters' weights and biases are randomly initialized.

After multiplying each filter with the input data, a non-linear activation function is added to the feature map to introduce complexity [41]. The mathematical formula for the CNN blocks is given by the following equations:

$$h_i = \text{ReLU}(w_i * h_{i-1} + b_i)$$
$$h_1 = \text{ReLU}(w_1 * x + b_1)$$
$$h_2 = \text{ReLU}(w_2 * h_1 + b_2)$$
$$h_3 = \text{ReLU}(w_3 * h_2 + b_3)$$

Where:

ReLU is a non-linear activation function.
$x$ is the input to the first convolutional layer.
$w_i$ is filters (kernels) of the $i$-th convolutional layer.
$b_i$ is the bias term of the $i$-th convolutional layer.
$h_1$, $h_2$, and $h_3$ are the hidden feature maps obtained after each convolutional layer.

After each CNN layer, a max-pooling layer is applied to generate a down-sampled version of the input feature map by choosing the maximum value of each feature within a specified area. This down-sampling process preserves the key spatial features while minimizing computational complexity and preventing overfitting [40]. The application of a max-pooling layer can be mathematically represented as:

$$p_i = MaxPooling(h_i)$$

Where:

$p_i$ is a down-sampled feature map after max-pooling.
$h_i$ is the feature map generated by the $i$-t convolutional layer.

The final feature map after the max-pooling process (p3) is reshaped from a 3-dimensional tensor into a one-dimensional vector. This flattening operation is an essential step for connecting the feature extraction layers to the fully connected layers in a neural network. The following is the mathematical representation of the flattening process:

$$f = flatten(p_3)$$

On the other hand, each LSTM block processes the input sequence step by step to extract temporal features. LSTM layers use gating mechanisms to learn long-term dependencies [37]. The following equations provide the mathematical formula for an LSTM block:

$$i_t = \sigma(w_i\, x_t + u_i h_{t-1} + b_i)$$
$$f_t = \sigma(w_f\, x_t + u_f h_{t-1} + b_f)$$
$$o_t = \sigma(w_o\, x_t + u_o h_{t-1} + b_o)$$
$$\tilde{c}_t = \tanh(w_c\, x_t + u_c h_{t-1} + b_c)$$
$$c_t = f_t \odot c_{t-1} + i_t \odot \tilde{c}_t$$

International Journal of Artificial Intelligence and Applications (IJAIA), Vol.16, No.5, September 2025

$$h_t = o_t \odot \tanh(c_t)$$
$$L = h_t$$

Where:

$i_t, f_t, o_t,$ are input, forget, and output gates.
$c_t$ is the cell state at time $t$.
$h_t$ is the hidden state at time $t$.
$\odot$ is element-wise multiplication.

$\sigma(z) = \frac{1}{1+e^{-z}}$ which is a sigmoid activation.

To create a unified representation that integrates both temporal and spatial features, the concatenation layer is responsible for combining the temporal feature from the LSTM ($L$) with $f$, which represents the final flattened feature map from the CNN [38]. The process can be represented mathematically as follows:

$$c = concatenate(L, f)$$

The combined outputs then pass through a fully connected layer to learn the relationships between the extracted features and map them to the final output, like class probabilities or regression values [16].

$$y = \text{ReLU}(wc + b)$$

Where:

ReLU is an activation function.
$c$ is the unified representation.
$w$ is the weight matrix of the fully connected layer.
$b$ is the bias vector of the fully connected layer.
After the fully connected layer, a dropout layer is applied to prevent overfitting by randomly disabling a fraction of neurons during training. The mathematical representation of the dropout operation is as follows:

$$d = dropout(y, m)$$

Where:
$y$ is the input to the dropout layer.
$m$ is the dropout rate.

Finally, the classification layer is connected to a Sigmoid layer to transform the output into a probability distribution between 0 and 1, allowing the classification layer to make accurate predictions about the labels [7].

The sigmoid function is defined as follows:

$$\sigma(z) = \frac{1}{1+\exp(-z)}$$

where:
*z* is the raw score (logit) for the classification layer.
*exp(-z)* is the exponential function applied to the negated logit.

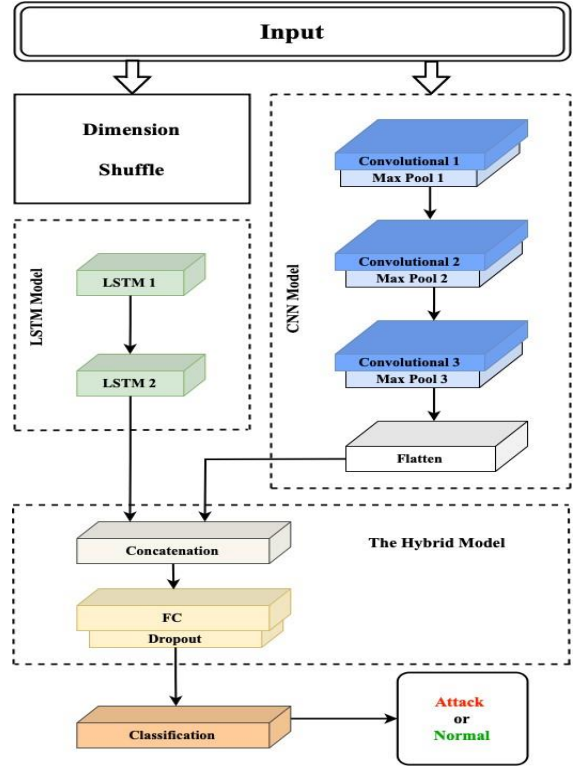

Figure 2. Proposed CNN–LSTM hybrid model

## 5. METHODOLOGY

The methodology employed in this research focuses on developing and evaluating a hybrid Convolutional Neural Network (CNN) and Long Short-Term Memory (LSTM) model for intrusion detection in smart grid communication systems. Two SCADA-based datasets, DNP3 and EIC 60870-5-104, were employed to reflect realistic traffic patterns in critical infrastructure. To ensure that the input features are consistent and suitable for deep learning, the data pre-processing stage handles missing values, normalization, and label encoding. The proposed model consists of three convolutional layers followed by pooling layers to extract spatial patterns, while the LSTM layers capture temporal dependencies across sequential traffic data. Hyperparameters such as learning rate, batch size, and dropout were fine-tuned using Bayesian optimization to achieve an optimal balance between performance and overfitting prevention. Model training was conducted using TensorFlow and the Adam optimizer with early stopping to prevent overfitting. The model's performance was assessed utilizing metrics such as accuracy, precision, recall, and F1-score.

## 5.1. Dataset Description and Preparation

DNP3 and EIC 60870-5-104 datasets are utilized by researchers to build intrusion detection systems based on DL techniques. By training deep learning models using these datasets, IDS can be developed to detect anomalies and potential cyber threats. Moreover, these datasets are particularly valuable for deep learning (DL) applications, providing well-structured and labeled flow statistics to train and evaluate models capable of detecting and mitigating various cyber threats in critical infrastructure systems.

### 5.1.1. DNP3 Intrusion Detection Dataset

Table 1. DNP3 samples for binary classification for training and testing the model.

| Type | Total Samples | Training Samples | Test Samples | Label |
| --- | --- | --- | --- | --- |
| Normal | 666 | 466 | 200 | 0 |
| Attack | 6,660 | 4,660 | 2000 | 1 |

Distributed Network Protocol 3 (DNP3) is a communication protocol that is primarily utilized in SCADA systems for industrial automation and control, particularly in the electricity and water utilities, such as smart grids. DNP3 enables remote communication between Industrial Control Systems (ICS) and SCADA, allowing for more efficient monitoring and management of critical infrastructure. DNP3 is a vital protocol designed to ensure that data is transmitted reliably, securely, and efficiently between control centers, Remote Terminal Units (RTUs), and Intelligent Electronic Devices (IEDs) [10].

The intrusion detection dataset (DNP3) [42] is publicly accessible with an extensive collection of network traffic data. DNP3 dataset includes normal activities and a wide range of malicious attacks within a SCADA environment. To reflect real-world threat scenarios, this dataset mainly focuses on cyberattacks such as DoS, Man-in-the-Middle (MitM), and unauthorized DNP3 commands injections, targeting DNP3 communication protocol. To enable advanced intrusion detection system (IDS) development, the DNP3 Intrusion Detection Dataset incorporates network traffic analysis based on key flow features such as timestamps, source and destination IPs, source and destination ports, protocols, and various cyberattacks. The dataset provides labeled flow statistics for both TCP/IP network traffic and DNP3-specific communication. While a labeled TCP/IP network flow statistics are generated using CICFlowMeter, a DNP3-specific flow statistics are provided by using a custom DNP3 Python parser to extract and analyze DNP3 communication packets within SCADA and ICS environments. These flow statistics are categorized based on the type of attack and are available for multiple flow timeout values (45, 60, 75, 90, 120, and 240 seconds). By offering realistic attack scenarios and extensive network flow data, the DNP3 dataset plays a crucial role in developing more resilient intrusion detection systems to enhance the security of ICS and SCADA networks against cyber threats.

### 5.1.2. EIC 60870-5-104 Intrusion Detection Dataset

Table 2. EIC 60870-5-104 Samples for Training and Testing the model for Binary Classification.

| Type | Total Samples | Training Samples | Test Samples | Label |
| --- | --- | --- | --- | --- |
| Normal | 569 | 400 | 169 | 0 |
| Attack | 6,259 | 4,400 | 1,859 | 1 |

IEC 60870-5-104 (IEC104) is a popular SCADA communication protocol utilized in smart grids and industrial automation for remote monitoring and control. IEC 60870-5-104 is an extension of IEC 608705-101 protocol with the development in transport, network, datalink, and physical layer services to facilitate real-time data exchange between control centers and remote terminal units (RTUs) at substations. Through TCP/IP networks, IEC 104 enables devices to transmit vital data such as voltage, current, and status updates. Even though IEC 104 improves operational efficiency, it raises several security challenges due to the lack of integrity and authentication [43]. Therefore, network traffic analysis is essential for security monitoring due to the vulnerability of IEC 104 to cyber threats like man-in-the-middle attacks, denial-of-service (DoS) attacks, and unauthorized data modification.

IEC 60870-5-104 dataset is demonstrated in [44] as an intrusion detection dataset which contain network traffic records of data transferred between devices utilizing the IEC 60870-5-104 protocol. The dataset was created using a network architecture of seven industrial entities executing IEC TestServer, one Human Machine Interfaces (HMI) utilized QTester104, and three attacking machines. Conversely, the cyberattacks employed Kali Linux integrated with Metasploit, OpenMUC j60870, and Ettercap. The dataset includes raw packet data obtained from equipment designed to monitor networks. These data packets include vital data such as source and destination IP addresses, source and destination ports, timestamps, and more. IEC 60870-5-104 dataset contains labeled network traffic that are categorized as normal and malicious activities with twelve IEC 60870-5-104 cyberattacks, including unauthorized commands and Denial of Service (DoS) attacks. Moreover, CICFlowMeter and a custom IEC 60870-5-104 Python Parser were employed to generate flow statistics by extracting features from the network flows. The custom parser produced 111 features, while CICFlowMeter extracted 83 features. These flow statistics are classified by attack type and are available in a variety of flow timeout values (15, 30, 60, 90, 120, and 180 seconds). Therefore, the IEC 60870-5-104 dataset is well-suited for training and enhancing robustness of AI-based intrusion detection models that utilize ML and DL techniques.

The first step in preparing the datasets for binary and multi-class classification using deep learning algorithms is data cleaning. Then the features in the datasets have been converted to numerical values and combined with other numerical features. For binary classification, as shown in tables 1 and 2, the labels in the datasets are numerically encoded by converting the "normal" labels to 0, which represent benign traffic, and converting the other cyberattack labels, such as MITM_DOS and STOP_APP, to 1. Furthermore, the datasets are min-max normalized to map all numerical features to a range of [0,1], aiming to decrease feature variance while maintaining feature associations. Tables 1 and 2 demonstrate that the datasets are divided into training and testing sets in a 70:30 ratio, with 70% assigned for model training and 30% assigned to validation and testing. In addition, while the training dataset enables the model to learn from the recognized behaviors of the network, the testing dataset can assess the model's capacity to identify unknown attacks. Eventually, the processed datasets are ready to be applied to train and test our deep learning model.

## 5.2. Performance Metrics

The performance of the hybrid model classifier is evaluated using various performance metrics derived from the confusion matrix. These metrics provide a comprehensive assessment of the model's effectiveness. The four key metrics that are utilized are represented mathematically as in [7] and written in subsequent equations as follows:

International Journal of Artificial Intelligence and Applications (IJAIA), Vol.16, No.5, September 2025

a. Classification Accuracy:

$$Accuracy\ (\%) = \frac{TP + TN}{TP + TN + FP + FN} \times 100$$

b. Precision:

$$Precision\ (\%) = \frac{TP}{TP + FP} \times 100$$

c. Recall:

$$Recall\ (\%) = \frac{TP}{TP + FN} \times 100$$

d. F1 Score:

$$F1\ score\ (\%) = 2 \times \frac{(precision \times recall)}{(precision + recall)} \times 100$$

Where:

TP (True Positive) indicates a correct prediction when the algorithm classifies an instance as positive, and it's truly positive.
TN (True Negative) indicates a correct prediction when the algorithm classifies an instance as negative, and it's truly negative.
FP (False Positive) indicates a wrong prediction when the algorithm classifies an instance as positive, but it is negative.
FN (False Negative) indicates a wrong prediction when the algorithm classifies an instance as negative, but it's positive.

**5.3. Implementation and Result Analysis**

To develop a reliable intrusion detection model, it's important to efficiently preprocess the datasets that are utilized to train, evaluate, and test the model. This involved data cleaning, normalization, and data encoding to ensure optimal input quality. After preprocessing the datasets, they have been applied to the proposed model. All experiments were executed in a Python environment using Jupyter Notebook, with Keras as the deep learning framework running on a CPU powered by an Apple M1 chip, 8 GB unified memory, and integrated 8-core Apple M1 GPU. We conducted extensive experimentation with several hyperparameter configurations to determine the optimal settings for enhancing model performance. The major goal of this hyperparameter modification was to improve the deep neural network's performance on the selected datasets.

Table 3. Hyperparameters for our model.

| Hyperparameter | Value |
|---|---|
| Learning rate | 0.001 |
| Epoch | 100 to 150 |
| Batch size | 16 |
| Optimizer | Adam |
| Dropout rate | 0.4 |
| Convolutional layers | 3 |
| LSTM units | 64 (1st layer), 128 (2nd layer) |

Hyperparameter tuning is crucial when developing a high-performing deep learning model, especially for complex tasks such as intrusion detection. The proposed CNN-LSTM architecture was fine-tuned through extensive experiments by focusing on hyperparameters that directly

International Journal of Artificial Intelligence and Applications (IJAIA), Vol.16, No.5, September 2025impact training dynamics and model generalization. Key hyperparameters that are shown in Table 3, including learning rate, batch size, number of epochs, dropout rate, and the depth of CNN and LSTM layers, were carefully tuned to obtain the best results. Training the model was stable and efficient by using the Adam optimizer (Adaptive Moment Estimation), which is known for its ability to adaptively adjust learning rates. Moreover, the learning rate is a key hyperparameter in training neural networks, which is responsible for controlling how much the model's weights are adjusted during each update step. For our model, a learning rate of 0.001 provided an ideal balance between convergence speed and stability, avoiding the risks of overshooting related to higher learning rates. Selecting an appropriate number of epochs is critical, since overtraining may lead the model to incorporate noise from the training data, resulting in worse performance on unseen data, while low epochs may cause the model to not train enough to learn patterns from the data. To avoid both overfitting and underfitting while ensuring sufficient training, the model was trained for 100 to 150 epochs with early stopping. Based on validation performance, the early stopping technique helps terminate training at the ideal time for generalization. Furthermore, a batch size of between 16 and 32 was chosen to achieve a balance between training efficiency and model accuracy, especially when dealing with limited GPU memory. While large batch sizes allow for capturing more global patterns and accelerating training, they also need more computational resources. In addition, dropout regularization was used at a rate of 0.4 to mitigate the problem of overfitting and improve generalization on unseen test data. An experimental evaluation of various dropout configurations revealed that this rate was the optimal value for our model and dataset, as it consistently reduced overfitting while maintaining high performance. Aiming to capture complex patterns while preserving computational efficiency, our proposed model comprises three convolutional layers for hierarchical feature extraction from the input data. Indeed, deeper convolutional networks can improve performance, but they also increase parameter count and resource consumption. Thus, three layers were chosen to achieve a balance of performance and resource efficiency. Subsequently, two layers of LSTM with 128 and 64 units are employed to capture the temporal relationships in network traffic data. This configuration was chosen to optimize performance while maintaining computational efficiency and mitigating overfitting.

Table 4. Performance Comparison of CNN-LSTM on DNP3 and IEC 60870-5-104 Datasets.

| Dataset | Accuracy | Precision | Recall | F1 Score |
|---|---|---|---|---|
| DNP3 | 99.68% | 99.69% | 99.95% | 99.82% |
| EIC 60870-5-104 | 99.70% | 99.84% | 99.72% | 99.78% |

Table 4 illustrates the performance of our proposed CNN-LSTM hybrid model on the DNP3 and IEC 608705-104 intrusion detection datasets to evaluate its effectiveness across different network protocols. To obtain the optimal results, we carefully modified hyperparameters such as batch size, epochs, and learning rate as described in Table 3. This fine-tuning process was essential to extract the most relevant features and improve the model's learning capacity and generalization performance. In terms of binary classification, our experimental results indicate that the proposed CNN-LSTM hybrid approach achieved excellent results with an accuracy of 99.68%, precision of 99.69%, recall of 99.95%, and an F1 score of 99.82% on the DNP3 dataset. In the IEC 60870-5-104 dataset, the model achieved an accuracy of 99.70%, a precision of 99.84%, a recall of 99.72%, and an F1 score of 99.78%, highlighting its strong performance across many data attributes and smart grid's network protocols. In addition, confusion matrix analysis for both datasets showed extremely low false positive and false negative rates, where high precision means very few false positives, and high recall means very few false negatives. These findings emphasize the model's robustness and efficiency, which demonstrate its competitive performance compared to other existing algorithms and its reliable capabilities to identify intrusions across many operational environments. For instance, the high recall value indicates that the model is highly efficient in accurately identifying both normal and invasive behaviors.

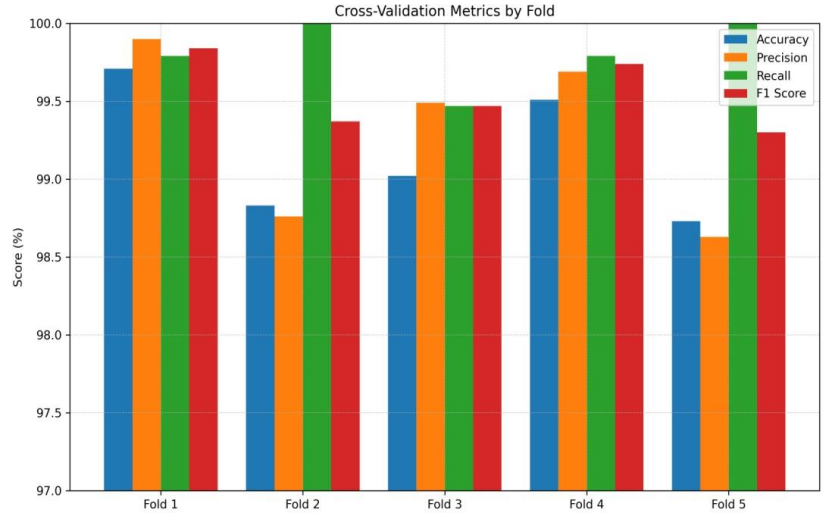

Figure 3. Cross-Validation Performance Metrics Across 5 Folds

To ensure the robustness and effectiveness of our proposed hybrid CNN-LSTM intrusion detection model, we used a statistical method called Stratified K-fold cross-validation. This method reduces overfitting and provides a statistically robust evaluation by employing stratified cross-validation that maintains class distributions over folds, which is crucial for accurate performance evaluation in intrusion detection scenarios. In addition, this rigorous process improves the scientific validity and repeatability of our conclusions compared to common evaluation techniques that depend on train-test splits. As shown in Figure 2, we utilized five-fold stratified cross-validation, which divided the dataset into five equal-sized subsets. In each iteration, one subset was selected as the validation set, while the remaining four were utilized for training. This process was repeated five times, with each subset performing once as a validation fold. Performance metrics such as accuracy, precision, recall, and F1-score were calculated for each fold, and the final performance values were the average of all folds. As a result, the model achieved an impressive average accuracy of 99.16%, precision of 99.29%, recall of 99.81%, and an F1 score of 99.54%, with a low average loss of 0.0630, indicating exceptional performance across all folds. Eventually, our use of stratified cross-validation introduces a reliable evaluation procedure that improves the scientific validity and reproducibility of the proposed IDS model, whereas many existing studies rely on fixed train-test splits to evaluate their IDS models. Our research enhances the current state of deep learning-based IDS methodologies by integrating this approach, thus offering a more rigorous, reliable, and fair evaluation framework for smart grid cybersecurity applications. In terms of intrusion detection, the proposed approach demonstrates distinctive performance compared to the existing standalone algorithms (see Table 5).

Table 5. Comparison with relevant works.

| Method | Accuracy | Precision | Recall | F1-score | Data | Year | Reference |
|---|---|---|---|---|---|---|---|
| ANN-ADS | 98.40% | 99.57% | 98.02% | 98.79% | CSE-CIC-IDS2018 | 2022 | [4] |
| CNN | 97.30% | 98.50% | 99.80% | 98.50% | Simulated data | 2023 | [34] |
| GRU | 98.60% | 99.50% | 97.40% | 98.50% | Simulated data | 2023 | [34] |
| LSTM | 87.15% | 97.88% | 76.83% | 86.09% | NSL-KDD | 2021 | [35] |
| XGB | 89.15% | 80.27% | 98.51% | 88.46% | NSL-KDD | 2021 | [35] |
| XGB-LSTM | 89.21% | 82.42% | 98.68% | 89.82% | NSL-KDD | 2021 | [35] |
| CNN–GRU–FL | 78.79% | 97.33% | 64.15% | 76.90% | NSL-KDD | 2023 | [36] |
| Decision Tree Classifier | 83.14% |  | 83.14% | 82.58% | IEC 60 870-5-104 | 2022 | [44] |
| Logistic Regression | 62.23% |  | 62.23% | 60.53% | IEC 60 870-5-104 | 2022 | [44] |
| Random Forest | 66.47% |  | 66.47% | 64.73% | IEC 60 870-5-104 | 2022 | [44] |
| Decision Tree | 93.05% | 93.06% | 93.05% | 93.05% | DNP3 | 2022 | [45] |
| DNN | 99.00% | 95.80% | 95.65% | 95.49% | DNP3 | 2022 | [45] |
| K-NN | 94.94% | 94.97% | 94.94% | 94.94% | DNP3 | 2022 | [45] |
| Naive Bayes | 68.27% | 72.22% | 64.90% | 68.27% | DNP3 | 2022 | [45] |
| Random Forest | 90.50% | 90.53% | 90.49% | 90.49% | DNP3 | 2022 | [45] |
| LSTM | 99.40% | 96.70% | 96.10% | 96.40% | CICIDS-2017 | 2023 | [46] |
| Proposed Algorithm | 99.70% | 99.84% | 99.72% | 99.78% | IEC 60 870-5-104 | 2025 | This Paper |
| Proposed Algorithm | 99.68% | 99.69% | 99.95% | 99.82% | DNP3 | 2025 | This Paper |

## 6. CONCLUSION AND FUTURE WORK

As electrical grids become more digitalized, it is critical to ensure the cybersecurity of smart grid communication networks. Protocols such as DNP3 and IEC 60870-5-104, which are essential components of SCADA-based infrastructures, provide real-time monitoring and control while also exposing the system to possible cyber threats such as unauthorized access and denial-of-service (DoS). In this research paper, we proposed a hybrid deep learning-based intrusion detection system (IDS) that combines CNN and LSTM networks to take advantage of their strengths in spatial features extraction and temporal patterns learning. The model was trained and validated on DNP3 and IEC 60870-5-104 intrusion detection datasets, which reflect real attack scenarios in smart grid environments. Through extensive experiment and hyperparameter adjustment, the hybrid CNN-LSTM model demonstrated improvements in the performance of detecting and classifying intrusions, obtaining a detection accuracy of 99.68% on the DNP3 dataset and 99.70% on the IEC 60870-5-104 dataset. Furthermore, high precision, recall, and F1 scores across both datasets prove the model's capability to accurately identify normal and malicious traffic on unseen data. Moreover, compared to existing intrusion detection techniques, our hybrid deep learning model outperformed in multiple evaluation metrics, illustrating the efficiency of hybrid deep learning architectures for intrusion detection systems in sophisticated smart grid environments.

In the future, we intend to improve the efficiency, scalability, and flexibility of the proposed CNN-LSTM hybrid model for smart grid intrusion detection. One important goal is to involve other deep learning architectures, such as transformers, GRUs, or attention mechanisms, to investigate their potential for capturing complex spatial and temporal data from network traffic. Furthermore, expanding the model's evaluation to a wide range of benchmark datasets and real-world smart grid scenarios will offer further insight about its robustness and generalizability. Another key focus will be on automating hyperparameter tuning using advanced optimization approaches, including Bayesian optimization and reinforcement learning-based strategies. This will significantly minimize human involvement while increasing model efficiency. Eventually, we intend to investigate a variety of optimization methodologies to improve the computational and memory efficiency of the proposed hybrid model while maintaining the intrusion detection system's accuracy and dependability.